\documentclass{IEEEtran}

\usepackage{etoolbox}
\makeatletter
\@ifundefined{color@begingroup}%
  {\let\color@begingroup\relax
   \let\color@endgroup\relax}{}%
\def\fix@ieeecolor@hbox#1{%
  \hbox{\color@begingroup#1\color@endgroup}}
\patchcmd\@makecaption{\hbox}{\fix@ieeecolor@hbox}{}{\FAILED}
\patchcmd\@makecaption{\hbox}{\fix@ieeecolor@hbox}{}{\FAILED}

\usepackage{cite}
\usepackage{amsmath,amssymb,amsfonts}
\usepackage{algorithmic}
\usepackage{lipsum}
\usepackage{graphicx}
\usepackage{textcomp}
\usepackage{multirow}
\usepackage{booktabs}
\usepackage{siunitx}
\usepackage{tabularx}
\usepackage{ragged2e}
\usepackage{etoolbox}   
\usepackage{nccmath}
\usepackage{url}

\DeclareMathOperator*{\argmin}{argmin}

\newcommand{\papertitle}{Optical Inversion and Spectral Unmixing of Spectroscopic Photoacoustic Images with Physics-Informed Neural Networks}

\newcolumntype{?}[1]{!{\vrule width #1}}

\newcolumntype{\?}{!{\vrule width 1pt}}

\renewcommand{\bfseries}{\fontseries{b}\selectfont} 
\robustify\bfseries 
\newrobustcmd{\B}{\bfseries}

\makeatletter
\newsavebox\myboxA
\newsavebox\myboxB
\newlength\mylenA

\newcommand*\xoverline[2][0.75]{%
    \sbox{\myboxA}{$\m@th#2$}%
    \setbox\myboxB\null% Phantom box
    \ht\myboxB=\ht\myboxA%
    \dp\myboxB=\dp\myboxA%
    \wd\myboxB=#1\wd\myboxA% Scale phantom
    \sbox\myboxB{$\m@th\overline{\copy\myboxB}$}%  Overlined phantom
    \setlength\mylenA{\the\wd\myboxA}%   calc width diff
    \addtolength\mylenA{-\the\wd\myboxB}%R
    \ifdim\wd\myboxB<\wd\myboxA%
       \rlap{\hskip 0.5\mylenA\usebox\myboxB}{\usebox\myboxA}%
    \else
        \hskip -0.5\mylenA\rlap{\usebox\myboxA}{\hskip 0.5\mylenA\usebox\myboxB}%
    \fi}
\def\BibTeX{{\rm B\kern-.05em{\sc i\kern-.025em b}\kern-.08em
    T\kern-.1667em\lower.7ex\hbox{E}\kern-.125emX}}
\def\ps@IEEEtitlepagestyle{%
  \def\@oddfoot{
      \scriptsize \parbox{\textwidth}{\hfil This work has been submitted to the IEEE for possible publication. Copyright may be transferred without notice, after which this version may no longer be accessible. \hfil}%
  }
}
\makeatother

\begin{document}
\title{\papertitle}
\author{Sarkis~Ter~Martirosyan,~Xinyue~Huang,~David~Qin,~Anthony~Yu,~and~Stanislav~Emelianov%
\thanks{This work was supported in part by the Breast Cancer Research Foundation (BCRF-22-043) and the National Institutes of Health (EB032822 and R01NS117613). In addition, Anthony Yu is funded in part by the US Army Medical Research and Development Command (Award W81XWH-19-1-0187). (Corresponding author: Stanislav Emelianov)}
\thanks{This work involved human subjects or animals in its research. All experiments involving animals were conducted with the approval of the Institutional Animal Care and Use Committee at the Georgia Institute of Technology.}
\thanks{Sarkis Ter Martirosyan was with the School of Electrical and Computer Engineering, Georgia Institute of Technology, Atlanta, GA 30332, United States while working on this project. He is now with the Institute of Biological and Medical Imaging, Helmholtz Zentrum Munich, 85764 Neuherberg, Germany as well as the Chair of Biological Imaging, Technische Universit\"at M\"unchen, 80333 M\"unchen, Germany. (e-mail: s.ter-martirosyan@tum.de).}
\thanks{Xinyue Huang, David Qin, and Anthony Yu are with The Wallace H. Coulter Department of Biomedical Engineering, Georgia Institute of Technology and Emory University School of Medicine, Atlanta, GA 30332, United States. (e-mails: xinyue@gatech.edu, dqin@gatech.edu, anthony.yu@gatech.edu)}
\thanks{Stanislav Emelianov is with the School of Electrical and Computer Engineering, Georgia Institute of Technology, Atlanta, GA 30332, United States as well as The Wallace H. Coulter Department of Biomedical Engineering, Georgia Institute of Technology and Emory University School of Medicine, Atlanta, GA 30332, United States (e-mail: stas@gatech.edu)}}

\maketitle

\begin{abstract}
Accurate estimation of the relative concentrations of chromophores in a spectroscopic photoacoustic (sPA) image can reveal immense structural, functional, and molecular information about physiological processes. However, due to nonlinearities and ill-posedness inherent to sPA imaging, concentration estimation is intractable. The Spectroscopic Photoacoustic Optical Inversion Autoencoder (SPOI-AE) aims to address the sPA optical inversion and spectral unmixing problems without assuming linearity. Herein, SPOI-AE was trained and tested on \textit{in vivo} mouse lymph node sPA images with unknown ground truth chromophore concentrations. SPOI-AE better reconstructs input sPA pixels than conventional algorithms while providing biologically coherent estimates for optical parameters, chromophore concentrations, and the percent oxygen saturation of tissue. SPOI-AE's unmixing accuracy was validated using a simulated mouse lymph node phantom ground truth. 
\end{abstract}


\section{Introduction}
\label{sec:introduction}

Spectroscopic Photoacoustic (sPA) imaging is a powerful medical imaging modality capable of revealing physiological information at centimeter depth with high spatial resolution \cite{emelianovPhotoacousticsMolecularImaging2009a, ntziachristosLookingListeningLight2005a, wangPracticalGuidePhotoacoustic2016a}. The contrast in a sPA image is derived from the presence of absorbing species, which can be either endogenous to the tissue or exogenous \cite{weberContrastAgentsMolecular2016a}. Quantifying the relative concentrations of endogenous or exogenous chromophores allows for clinical insights to be gleaned \cite{liPhotoacousticTomographyBlood2018a}. By determining the relative concentrations of oxygenated hemoglobin (HbO2) and deoxygenated hemoglobin (HHb), a spatial map of oxygen saturation (SO2) in tissue can be found. Computing such an SO2 map can, for example, track hypoxia in vasculature to identify malignant tumors \cite{kubelick2026ultrasound}, visualize the functional parameters of traumatic brain injuries, and study neuronal activities \cite{wangNoninvasiveImagingHemoglobin2006a}.

To calculate an SO2 map correctly, it is vital that the relative concentrations of HbO2 and HHb are estimated accurately. Oxygen saturation estimates must be accurate since even small errors can confound diagnoses \cite{qin2025depth}. In the case of identifying malignant tumors based on the oxygen saturation of vasculature, the difference between normal and abnormal SO2 values can be as little as 10 percent \cite{mallidiPhotoacousticImagingCancer2011a, vaupelTumorHypoxiaMalignant2004a}. Moreover, the difference between normoxia and hyperoxia in rat cerebral vasculature can be as little as nine percent \cite{wangNoninvasiveImagingHemoglobin2006a}. Improved SO2 estimation can allow for more effective imaging of skin \cite{huang2025ultrasound, long2025high}, breast tissue, and the brain \cite{boasHaemoglobinOxygenSaturation2011a}.

Based on these considerations, it is paramount that the relative chromophore concentration estimation algorithm---also known as the spectral unmixing algorithm---yields accurate results. The approaches discussed herein can be separated into two categories: linear and nonlinear. Linear spectral unmixing algorithms assume that a sPA image is a linear combination of chromophore concentrations \cite{lukeOpticalWavelengthSelection2013a}. Conversely, nonlinear algorithms try to compensate for wavelength-dependant fluence attenuation and optical scattering \cite{yuanQuantitativePhotoacousticTomography2006a}. 

The linear unmixing algorithms explored herein are nonnegative-least-squares (NLS) and nonnegative-matrix-factorization (NMF). NLS spectral unmixing uses pure chromophore absorption spectra taken from literature to estimate relative absorption concentrations and is widely used thanks to its straightforward implementation \cite{benchQuantitativePhotoacousticEstimates2021a}. NMF spectral unmixing, unlike NLS, leverages a data-driven mechanism to better describe sPA data \cite{grassoAutomaticUnmixingApproach2020b}. There are other data-driven linear spectral unmixing algorithms available, notably principal-component-analysis and independent-component-analysis \cite{glatzBlindSourceUnmixing2011b}. However, it was shown in \cite{grassoRecentAdvancesPhotoacoustic2022a} that NMF outperforms the alternative data-driven methods when unmixing sPA images, especially in tissues with high background absorption. 

There are many nonlinear spectral unmixing approaches designed to better estimate relative chromophore concentrations. Some nonlinear approaches leverage simplifying assumptions to estimate the optical fluence of tissue. One notable example of such an algorithm is eigenspectra-multispectral-optoacoustic-tomography (eMSOT) \cite{tzoumasEigenspectraOptoacousticTomography2016a}. The eMSOT method assumes that optical fluence can be estimated by a linear combination of so-called eigenspectra extracted using a principal-component analysis of simulated fluence spectra. 

However, most nonlinear spectral unmixing approaches opt to leverage machine learning and neural networks. Several approaches estimate SO2 directly rather than spectrally unmixing first. The learned-spectral-decoloring (LSD) method and convolutional-encoder–decoder-with-skip-connections (EDS) both accurately estimate the oxygen saturation of \textit{in silico} sPA images \cite{grohlLearnedSpectralDecoloring2021a, benchAccurateQuantitativePhotoacoustic2020a}. The EDS method also uses 3D-convolutional-neural-networks (3D-CNNs) to incorporate spatial relationships into the estimation mechanism. 

Another approach using CNNs is the quantitative-optoacoustic-tomography-network (QOAT-Net) \cite{liDeepLearningbasedQuantitative2022b}. QOAT-Net uses parallel U-Nets to estimate the optical fluence and absorption coefficients of a photoacoustic (PA) image. Similarly to the aforementioned machine learning unmixing methods, QOAT-Net was trained using \textit{in silico} PA images with known fluence and absorption. However, the QOAT-Net project made use of a generative-adverserial-network (GAN) to modify \textit{in silico} images so that those images were indiscriminable from \textit{in vivo} ones. This allowed for QOAT-Net to be trained fully-supervised while better fitting \textit{in vivo} sPA images.  

It is also possible to design a machine-learning based spectral unmixing method trained on labeled phantom images. In \cite{grohlMovingSimulationDataDriven2024}, a U-Net (called ``DL-Exp'') was used to estimate absorption coefficients from spectroscopic photoacoustic images. DL-Exp was trained on sPA images of well-characterized mineral oil phantoms with varied nigrosin concentrations to set absorption properties and titanium oxide concentrations to set scattering properties. By carefully measuring the optical properties of the phantoms used for training, DL-Exp can estimate the absorption coefficients as well as the SO2 of \textit{in vivo} mouse sPA images.

While the various nonlinear methods successfully unmix photoacoustic images after being trained on labeled \textit{in silico} images, none provide a framework for semi-supervised or self-supervised learning. In other words, none of the nonlinear approaches can directly replace blind linear approaches like NLS and NMF. A deep learning method that is able to blindly unmix \textit{in vivo} sPA images without relying on ground truth data for training while simultaneously incorporating fluence correction would be a valuable tool for photoacoustic optical inversion. 

In this work, we propose the spectroscopic-photoacoustic-optical-inversion-autoencoder (SPOI-AE), a deep autoencoder architecture trained to optically invert and spectrally unmix \textit{in vivo} sPA pixels. SPOI-AE was trained in a self-supervised manner on sPA pixels from images of mouse lymph \cite{yu2023development}. The SPOI-AE utilizes a deterministic approximate model for optical transport in diffuse media as its decoder stage; from a physics-informed deep learning perspective, this embeds knowledge of optical transport into the loss calculation, heavily penalizing spectral unmixing and optical inversion results that violate our understanding of optical transport in mouse sPA imaging. Our neural network design and training strategy comes from how physics-informed neural networks integrate governing physics into training to improve generalization with limited data \cite{karniadakisPhysicsinformedMachineLearning2021}. We demonstrate that SPOI-AE can optically invert and spectrally unmix \textit{in vivo} sPA images more effectively than the alternative self-supervised methods by incorporating optical inversion and fluence compensation.  

% \clearpage
\section{Methods}

\subsection{Optical Forward Problem}
\label{sec:fwd}

The photoacoustic effect states that when tissue is illuminated by short laser pulses, energy is absorbed by the tissue, causing thermoelastic expansion and acoustic propagation \cite{rosencwaigTheoryPhotoacousticEffect1976a}. The acoustic waves triggered by thermoelastic expansion can be measured by an ultrasound transducer, creating a photoacoustic image. A spectroscopic photoacoustic image is formed by collating multiple photoacoustic images captured at different laser wavelengths \cite{lukeOpticalWavelengthSelection2013a}. A sPA image is indexed with respect to spatial displacement from the transducer and laser wavelength---denoted by $\mathbf{r}$ and $\lambda$, respectively. A sPA image $p(\mathbf{r}, \lambda)$ is a function of the tissue's absorption coefficient $\mu_a(\mathbf{r}, \lambda)$ and the reduced scattering coefficient $\mu_s'(\mathbf{r}, \lambda)$\cite{coxQuantitativeSpectroscopicPhotoacoustic2012a}. This relationship is known as the photoacoustic optical forward problem and is described as

\begin{equation}
\label{eq:fwd}
p(\mathbf{r}, \lambda) = \Gamma \Phi\!\left(\mathbf{r}, \lambda; \mu_a, \mu_s'\right) \mu_a\!\left(\mathbf{r}, \lambda\right)
\end{equation}

\noindent where $\Gamma$ is the Gr\"uneisen coefficient and $\Phi$ is the optical fluence. The optical fluence is itself a function of $\mu_a$ and $\mu_s'$ based on the diffusion approximation in homogeneous media

\begin{equation}
\label{eq:fluence}
\begin{aligned}
\mu_{\mathrm{eff}}(\mathbf{r}, \lambda) &= \sqrt{3\mu_a(\mathbf{r}, \lambda)\left(\mu_a(\mathbf{r}, \lambda) + \mu_s'(\mathbf{r}, \lambda)\right)}\\
\Phi\!\left(\mathbf{r}, \lambda\right) &= \phi_0(\lambda) \exp \left\{-  \mu_{\mathrm{eff}}(\mathbf{r}, \lambda)\left|\mathbf{r}\right|\right\}
\end{aligned}
\end{equation}

\noindent where $\mu_{\mathrm{eff}}(\mathbf{r}, \lambda)$ denotes the effective attenuation coefficient which summarizes the effects of absorption and scattering. Moreover, the variable $\phi_0(\lambda)$ denotes  wavelength-dependant fluence attenuation which varies only with wavelength, not spatial position. The absorption coefficient $\mu_a(\mathbf{r}, \lambda)$ is a conical combination of $N$ absorption spectra $\varepsilon_n(\lambda) \in \mathbb{R}_{\geq 0},\,n=1,\ldots,N$, described as

\begin{equation}
\label{eq:mua}
\mu_a(\mathbf{r}, \lambda) = \sum_{n=1}^N \varepsilon_n(\lambda) c_n(\mathbf{r}).
\end{equation}

The absorption spectra relate how relative chromophore concentrations $c_n(\mathbf{r}) \in \mathbb{R}_{\geq 0}$ affect the absorption coefficient of the tissue. Absorption spectra can describe both endogenous species like HbO2 and HHb as well as exogenous chromophores like gold nanostructures and small-molecule dyes \cite{fuPhotoacousticImagingContrast2019a}.

\subsection{Optical Inverse Problem}

The optical forward problem provides a procedure for how to compute a sPA image given the optical parameters. However, reversing this process to estimate $\mu_a(\mathbf{r}, \lambda)$ and $\mu_s'(\mathbf{r}, \lambda)$ from a sPA image is highly challenging. This problem, known as the photoacoustic optical inverse problem, is intractable as it is nonlinear and ill-posed \cite{coxQuantitativeSpectroscopicPhotoacoustic2012a}. Fortunately, deep learning provides a powerful framework for solving nonlinear and ill-posed problems in medical imaging \cite{adlerSolvingIllposedInverse2017a}. Furthermore, deep autoencoders can be used with a deterministic decoding stage to accurately estimate a physically coherent latent space \cite{zhaoHyperspectralUnmixingAdditive2022a}. With these properties in mind, a deep autoencoder architecture is a promising method for solving the photoacoustic optical inverse problem. 

% \brown{This paragraph might be better in \textbf{Introduction}, see comments.} \red{I think it is important to have this section to describe optical inversion in terms of the ``mathematical'' terms established in \S \ref{sec:fwd} }

% To me, this paragraph is more like an introduction (of why we would like to use deep learning) instead of specific methods.

% You could remove this subsection; move the "Fortunately, deep learning..." part to the second to last paragraph of the <Introduction> section; add one sentence about optical inverse problem at the end of the <Optical Forward Problem> subsection, and may change its name to <Optical Forward and Inverse Problem>.
% Only my opinion, we might discuss about it together.

\subsection{Linear Spectral Unmixing}
\label{sec:linsu}

Rather than estimating a solution the optical inverse problem, it is common to assume a linear relationship between a sPA signal and the absorption coefficient, i.e. that $p(\mathbf{r}, \lambda) \approx \mu_a(\mathbf{r}, \lambda)$. With this assumption, estimating chromophore concentrations only involves an inverse problem with respect to \eqref{eq:mua}. The two methods for inverting \eqref{eq:mua} explored herein are nonnegative-least-squares with literature spectra (Lit. NLS) and nonnegative-matrix-factorization (NMF). 

Nonnegative-least-squares with literature spectra defines the pure-chromophore absorption spectra $\varepsilon_n(\lambda_l)$ based on values tabulated from literature, for example from \cite{prahlTabulatedMolarExtinction1998a}. By defining $\mathbf{E} \in \mathbb{R}_{\geq 0}^{L \times N}$ such that $\varepsilon_{ln} = \varepsilon_n(\lambda_l)$ and $\mathbf{c}(\mathbf{r}_i) \in \mathbb{R}_{\geq 0}^{N}$ such that $c_n(\mathbf{r}_i)$ is the relative concentration of the $n$th chromophore, \eqref{eq:mua} can be vectorized at a spatial location $\mathbf{r}_i$

\begin{equation}
\label{eq:muavec}
\boldsymbol{\mu}_a(\mathbf{r}_i) = \mathbf{E} \mathbf{c}(\mathbf{r}_i). 
\end{equation}

\noindent The Lit. NLS inversion can be written in terms of this vectorized reformulation as 

\begin{equation}
\label{eq:litnls}
\widehat{\mathbf{c}}(\mathbf{r}_i) = \argmin_{\mathbf{c} \in \mathbb{R}_{\geq 0}^{N}}   \left\| \mathbf{E} \mathbf{c} - \mathbf{p}(\mathbf{r}_i) \right\|_2^2
\end{equation}

\noindent where $\widehat{\mathbf{c}}(\mathbf{r}_i)$ is the vector of estimated chromophore concentrations. 

It is also possible to leverage a data-driven approach to inverting \eqref{eq:mua}, where the absorption spectra are tuned to fit the data. In this study, the data-driven approach explored is nonnegative-matrix-factorization, which is a proven sPA unmixing method. Using the vectorized representation \eqref{eq:muavec}, NMF can be written as 

\begin{equation}
\label{eq:nmf}
\widehat{\mathbf{C}}, \widehat{\mathbf{E}} = \argmin_{\mathbf{C} \in \mathbb{R}_{\geq 0}^{I \times N},\, \mathbf{E}\in\mathbb{R}_{\geq 0}^{L \times N}}   \left\| \mathbf{C}\mathbf{E}^T - \mathbf{P} \right\|_2^2
\end{equation}

\noindent where $I$ is the number of pixels in the sPA dataset, $\widehat{\mathbf{C}} \in \mathbb{R}_{\geq 0}^{I \times N}$ is the matrix of estimated chromophore concentrations, $\widehat{\mathbf{E}} \in \mathbb{R}_{\geq 0}^{N \times L}$ is the matrix of estimated absorption spectra, and $\mathbf{P} \in \mathbb{R}_{\geq 0}^{I \times L}$ is the sPA design matrix. $\mathbf{P}$ is set up so that $p_{il} = p(\mathbf{r}_i, \lambda_l)$ where $\mathbf{r}_i$ is the displacement for the $i$th pixel and $\lambda_l$ is the $l$th laser wavelength.

\subsection{SPOI-AE Architecture}
\label{sec:aearch}

The goal of SPOI-AE is to estimate the optical parameters and relative chromophore concentrations of an arbitrary sPA pixel. From an input sPA pixel $p(\mathbf{r}, \lambda)$, SPOI-AE is able to estimate optical parameters and relative chromophore concentrations---the latter representing SPOI-AE's latent space. The absorption coefficient $\mu_a(\mathbf{r}, \lambda)$ is estimated via a fully-connected neural network (FCNN) called ``$\mu_a$-Net'' and the reduced scattering coefficient  $\mu_s'(\mathbf{r}, \lambda)$ is similarly estimated by another FCNN called ``$\mu_s'$-Net.'' The input sPA pixel is then recreated from the FCNN generated optical parameter estimates using the machinery introduced in Section \ref{sec:fwd}. Simultaneously, the relative chromophore concentrations are unmixed from the absorption coefficient estimate. SPOI-AE's architecture---shown graphically in Fig. 1---allows for optical inversion to be learned in a self-supervised manner, thanks to the aforementioned deterministic decoding stage.

% \brown{Confusing wording regarding what is SPOI-AE's ``output.''} \red{Better now? I tried to better emphasize the difference between the ``reconstruction output'' and the latent space. We are interested in the latent space, but need the reconstruction for self-supervised learning. }

\subsection{Neural Network Design}
\label{sec:nndesign}

The FCNN structure is a composition of batch-normalized (BN) fully-connected layers activated by the leaky rectified linear unit (LReLU) activation function. Each FCNN is terminated by a rectified linear unit (ReLU) activation function. The LReLU, ReLU, and BN funtions are defined as

\begin{equation}
\mathrm{LReLU}(\mathbf{X}) = \begin{cases} x_{im} & x_{im} \geq 0 \\ 0.01 x_{im} & x_{im} < 0 \end{cases}\ \forall\ i,\,m
\end{equation}

\begin{equation}
\mathrm{ReLU}(\mathbf{X}) = \begin{cases} x_{im} & x_{im} \geq 0 \\ 0 & x_{im} < 0 \end{cases}\ \forall\ i,\,m
\end{equation}

\begin{equation}
\mathrm{BN}(\mathbf{X}) = \frac{\mathbf{X} - \mathrm{E}\left[\mathbf{X}\right]}{\sqrt{\mathrm{Var}\left[\mathbf{X}\right] + \texttt{EPSILON}}}
\end{equation}

\noindent where the input $\mathbf{X}$ is a two-dimensional matrix such that $\mathbf{X} \in \mathbb{R}^{I \times M_j}$. The dimensional constants are $I$ and $M_j$,  the batch-size and the number of features in the current layer $j$, respectively. The operator $\mathrm{E}[\cdot]$ denotes the expected value and $\mathrm{Var}[\cdot]$ the variance. The term $\texttt{EPSILON}$ represents a very small constant that ensures numerical stability. 

The building blocks within the FCNN are defined as

\begin{equation}
\label{eq:layer}
\begin{aligned}
f_j(\mathbf{X}) &= \boldsymbol\gamma_j \odot \mathrm{BN}\left(\mathrm{LReLU}\left(\mathbf{X}\mathbf{W}_j^T + \mathbf{b}_j\right) + \boldsymbol\beta_j\right) 
\\
g_j(\mathbf{X}) &= \mathrm{ReLU}\left(\mathbf{X} \mathbf{W}_j^T + \mathbf{b}_j\right)
\end{aligned}
\end{equation}

\noindent where $\mathbf{W}_j \in \mathbb{R}^{M_j \times M_{j-1}}$ and $\mathbf{b}_j \in \mathbb{R}^{M_j}$ are the set of learned weights and learned biases for the $j$th layer, respectively. The terms $\boldsymbol\gamma_j \in \mathbb{R}^{1 \times M_j}$ and $\boldsymbol\beta_j \in \mathbb{R}^{1 \times M_j}$ are the affine parameters of the batch-normalization for the $j$th layer. The operator $\odot$ is the element-wise matrix product, also known as the Hadamard product. 

Finally, the FCNN architecture is described analytically as

\begin{equation}
\label{eq:nets}
\begin{aligned}
\mathbf{M}_a(\mathbf{P}) &= g_{3\mathrm{a}} \circ f_{2\mathrm{a}} \circ f_{1\mathrm{a}}(\mathbf{P})
\\
\mathbf{M}_s'(\mathbf{P }) &= g_{4\mathrm{s}} \circ f_{3\mathrm{s}} \circ f_{2\mathrm{s}} \circ f_{1\mathrm{s}}(\mathbf{P})
\end{aligned}
\end{equation}

\noindent where the building blocks described in \eqref{eq:layer} are used to define $\mu_a$-Net and $\mu_s'$-Net. The additional subscripts ``$\mathrm{a}$'' and ``$\mathrm{s}$'' are used to differentiate  terms having the same number subscript (e.g., between $f_{1\mathrm{a}}$ and $f_{1\mathrm{s}}$). The input is the same sPA design matrix $\mathbf{P}$ used with NMF in \eqref{eq:nmf}. The nonnegative output matrices $\mathbf{M}_a(\mathbf{P}) \in \mathbb{R}^{I \times L}_{\geq 0}$ and $\mathbf{M}_s'(\mathbf{P}) \in \mathbb{R}^{I \times L}_{\geq 0}$ are calculated as a function of $\mathbf{P}$ as well as the various learned parameters, which are set up so that $\mu_{a,il} = \mu_a(\mathbf{r}_i, \lambda_l)$ and $\mu_{s,il}' = \mu_s'(\mathbf{r}_i, \lambda_l)$. 

\begin{figure*}[htb]
    \centering
    \includegraphics[width=\textwidth]{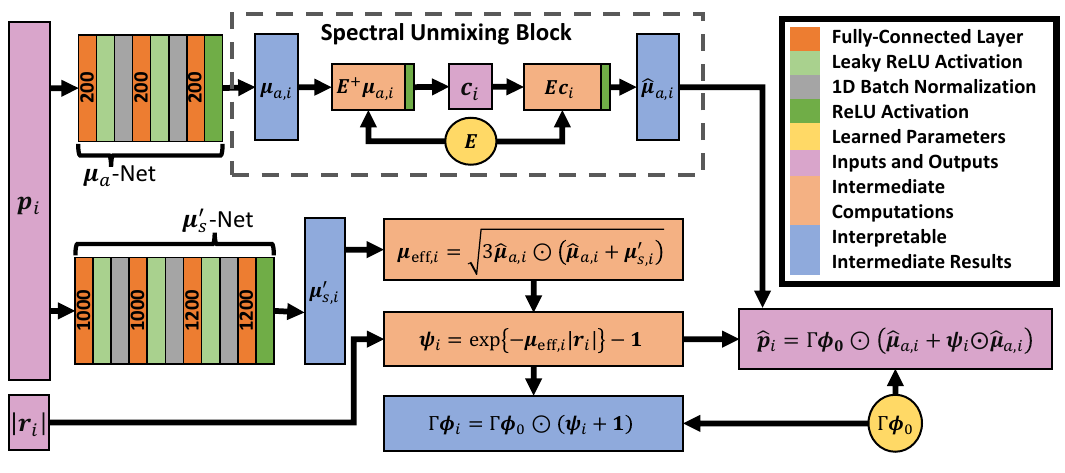}
    \caption{The complete architecture of SPOI-AE. For each sPA pixel numbered $i$, the inputs are $\mathbf{p}_i$ and $|\mathbf{r}_i|$, the input pixel and the pixel's distance from the origin, respectively. The reconstructed output pixel is denoted with $\widehat{\mathbf{p}}_i$. The FCNNs $\boldsymbol{\mu}_s'$-Net and $\boldsymbol{\mu}_a$-Net generate estimates for the absorption coefficient $\boldsymbol{\mu}_{a,i}$ and the reduced scattering coefficient $\boldsymbol{\mu}_{s,i}'$. The spectral unmixing block produces chromophore concentration estimates $\boldsymbol{c}_i$ from $\boldsymbol{\mu}_{a,i}$. A low-rank reconstruction of $\boldsymbol{\mu}_{a,i}$ is recreated from the chromophore concentration estimates denoted as $\widehat{\boldsymbol{\mu}}_{a,i}$. Finally, the decoding stage implementing the photoacoustic optical forward problem described in section \ref{sec:fwd}, reconstructs $\widehat{\mathbf{p}}_i$ from $\widehat{\boldsymbol{\mu}}_a$ and $\boldsymbol{\mu}_s'$. }
    \label{fig:deepae}
\end{figure*}

\subsection{Latent Representation and Spectral Unmixing}
\label{sec:latent}

The FCNNs described in Section \ref{sec:nndesign} allow us to estimate the optical parameters $\mu_a$ and $\mu_s'$, which are the primary latent variables of the deep autoencoder design. However, as indicated in \eqref{eq:mua}, the absorption coefficient can be further decomposed with respect to the absorption spectra to calculate the relative chromophore concentrations.

\noindent If the absorption spectra are known, it is possible to invert \eqref{eq:muavec} via rectified least squares

\begin{equation}
\label{eq:cvec}
\widehat{\mathbf{c}}(\mathbf{r}_i) = \mathrm{ReLU}\!\left(\mathbf{E}^+\boldsymbol{\mu}_a(\mathbf{r}_i)\right)
\end{equation}

\noindent where $\mathbf{E}^+$ denotes the Moore-Penrose pseudoinverse of $\mathbf{E}$. Finally, by applying \eqref{eq:muavec} to the unmixing concentrations found using \eqref{eq:cvec}, a rank-$N$ approximation of $\mu_a$ can be found

\begin{align}
\begin{split}
\label{eq:muahat}
\widehat{\boldsymbol{\mu}}_a(\mathbf{r}_i) &= \mathrm{ReLU}\!\left(\mathbf{E}\,\widehat{\mathbf{c}}(\mathbf{r}_i)\right) \\
&= \mathrm{ReLU}\!\left(\mathbf{E}\,\mathrm{ReLU}\!\left(\mathbf{E}^+\boldsymbol{\mu}_a(\mathbf{r}_i)\right)\right). 
\end{split}
\end{align}

Performing spectral unmixing given absorption coefficients is often easy in practice thanks to tabulated absorption spectra from literature. For example, \cite{prahlTabulatedMolarExtinction1998a} provides experimentally derived absorption spectra for HbO2 and HHb. However, the SPOI-AE deep learning framework can adjust absorption spectra from literature as a byproduct of the machine learning training process. This can potentially compensate for residual nonlinearities related to optical scattering or spectral coloring not isolated by $\mu_s'$-Net \cite{lauferVitroMeasurementsAbsolute2005a, hochuliEstimatingBloodOxygenation2019a}.

\subsection{Deterministic Decoding of the Latent Space}

As indicated in Fig. \ref{fig:deepae} and Section \ref{sec:aearch}, the decoding stage of SPOI-AE involves reconstructing the input sPA pixel from the latent variables $\widehat{\mu}_a$ and $\mu_s'$. Reconstructing the relevant sPA input pixel from its latent representation is a deterministic process, as we are able to leverage the photoacoustic optical forward problem introduced in section \ref{sec:fwd}. The optical forward problem is vectorized as 

\begin{equation}
\label{eq:fwdvec}
\begin{aligned}
\widehat{\mathbf{C}}(\mathbf{P}) &= \left(\mathrm{ReLU}\!\left(\mathbf{E}^+\mathbf{M}_a^T\!(\mathbf{P})\right)\right)^T
\\
\widehat{\mathbf{M}}_a(\mathbf{P}) &= \left(\mathrm{ReLU}\!\left(\mathbf{E}\widehat{\mathbf{C}}^T(\mathbf{P})\right)\right)^T
\\
\mathbf{M}_\mathrm{eff}(\mathbf{P}) &= \sqrt{3\widehat{\mathbf{M}}_a(\mathbf{P}) \odot \left(\widehat{\mathbf{M}}_a(\mathbf{P}) + \mathbf{M}_s'(\mathbf{P})\right)}
\\
\boldsymbol{\Phi}(\mathbf{P}, \boldsymbol{\rho}) &= \boldsymbol{\phi}_0 \odot \exp \left\{-\mathbf{M}_\mathrm{eff}(\mathbf{P}) \odot \boldsymbol{\rho} \right\}
\\
\widehat{\mathbf{P}}(\mathbf{P}, \boldsymbol{\rho}) &= \Gamma \boldsymbol{\Phi}(\mathbf{P}, \boldsymbol{\rho}) \odot \widehat{\mathbf{M}}_a(\mathbf{P})
\end{aligned}
\end{equation}

\noindent where $\widehat{\mathbf{C}}(\mathbf{P})$ represents the matrix of relative concentrations for the input design matrix $\mathbf{P}$ such that $\widehat{\mathbf{C}}(\mathbf{P}) \in \mathbb{R}_{\geq 0}^{I \times N}$ and $\widehat{c}_{in}(\mathbf{P})$ is the relative concentration of the $n$th absorbing species at pixel $i$, and $\widehat{\mathbf{M}}_a(\mathbf{P})$ is the rank-$N$ reconstruction of the estimated absorption coefficients $\mathbf{M}_a(\mathbf{P})$ as defined in  \eqref{eq:muahat}. $\mathbf{M}_\mathrm{eff}(\mathbf{P})$ and $\boldsymbol{\Phi}(\mathbf{P}, \boldsymbol{\rho})$ follow from \eqref{eq:fluence} and are the estimated effective attenuation coefficient matrix and estimated optical fluence matrix, respectively. The term $\widehat{\mathbf{P}}(\mathbf{P}, \boldsymbol{\rho})$ is the reconstruction of the input design matrix $\mathbf{P}$ based on \eqref{eq:fwd}. The vector $\boldsymbol{\rho}$ is defined such that $\boldsymbol{\rho} \in \mathbb{R}_{\geq 0}^{I}$ and $\rho_i = |\mathbf{r}_i|$. Finally, the term $\boldsymbol{\phi}_0$ is the initial fluence at surface vector such that $\boldsymbol{\phi}_0 \in \mathbb{R}_{\geq 0}^{1 \times L}$ and $\phi_{0,l} = \phi_0(\lambda_l)$ where $\lambda_l$ is the $l$th laser wavelength. 

For performance considerations, the vectorized forward problem in \eqref{eq:fwdvec} was modified to be the sum of a linear portion dependant only on $\hat{\mu}_a$ and a nonlinear portion. This sort of decomposition allows for SPOI-AE to learn the latent space more effectively \cite{zhaoHyperspectralUnmixingAdditive2022a}.  This decomposition is derived as

\begin{equation}
\label{eq:fwddecomp}
\begin{aligned}
\widehat{\mathbf{P}}(\mathbf{P}, \boldsymbol{\rho}) &= \Gamma \boldsymbol{\Phi}(\mathbf{P}, \boldsymbol{\rho}) \odot \widehat{\mathbf{M}}_a(\mathbf{P}) 
\\
&= \Gamma \boldsymbol{\phi}_0 \odot \exp \left\{-\mathbf{M}_\mathrm{eff}(\mathbf{P}) \odot \boldsymbol{\rho} \right\} \odot \widehat{\mathbf{M}}_a(\mathbf{P}) 
\\
&= \Gamma \boldsymbol{\phi}_0 
\\&\quad \odot \left(\exp \left\{-\mathbf{M}_\mathrm{eff}(\mathbf{P}) \odot \boldsymbol{\rho} \right\} - \mathbf{1} + \mathbf{1}\right) 
\\&\quad \odot \widehat{\mathbf{M}}_a(\mathbf{P}) 
\\
&= \Gamma \boldsymbol{\phi}_0 \odot \left(\exp \left\{-\mathbf{M}_\mathrm{eff}(\mathbf{P}) \odot \boldsymbol{\rho} \right\} - \mathbf{1}\right) 
\\&\qquad \odot \widehat{\mathbf{M}}_a(\mathbf{P}) 
\\&\quad + \Gamma \boldsymbol{\phi}_0 \odot \widehat{\mathbf{M}}_a(\mathbf{P}) 
\\
&= \Gamma \boldsymbol{\phi}_0 \odot \left(\boldsymbol{\Psi}(\mathbf{P}, \boldsymbol{\rho}) \odot \widehat{\mathbf{M}}_a(\mathbf{P})  + \widehat{\mathbf{M}}_a(\mathbf{P})\right)
\end{aligned}
\end{equation}

\noindent where the term $\boldsymbol{\Psi}\!\left(\mathbf{P}, \boldsymbol{\rho}\right)$ is the nonlinear multiplier such that $\boldsymbol{\Psi}\!\left(\mathbf{P}, \boldsymbol{\rho}\right) \in \left(-1,0\right]^{I \times L}$. The decomposition above reveals SPOI-AE's final learnable parameter: the combination of the Gr\"uneisen coefficient $\Gamma$ and the wavelength-dependant fluence $\boldsymbol{\phi}_0$, denoted by $\Gamma \boldsymbol{\phi}_0$. Unfortunately, without prior knowledge of $\Gamma$, it is impossible to separate these terms. If $\Gamma$ is found by a separate process, it would be possible to extract $\boldsymbol{\phi}_0$. \\

% \brown{Starting from here can be \textbf{Experiments}, but keep it within \textbf{Methods} is fine! Also, I think we could move the \textbf{Section G} after the \textbf{Sction H} and \textbf{I}.} \red{Done. Opted to keep experiments in Methods but added an Evaluation Metrics section.  }

\subsection{Dataset Composition}

The dataset used for training and testing was composed of \textit{in vivo} spectroscopic photoacoustic images of subiliac mouse lymph nodes. These images were captured using the
VisualSonics VevoLAZR system (FujiFilm VisualSonics Inc., Toronto, Canada) and the LZ-550 transducer (256 elements, f\textsubscript{c} = 40 MHz) with L = 146 laser wavelengths ranging from 680 nm to 970 nm at 2 nm intervals—in the so-called “near-infrared window," no frame averaging was used. All together eleven sPA images were used for training and three for testing, collected from seven mice.  

\begin{figure}[htb]
    \centerline{\includegraphics[width=\columnwidth]{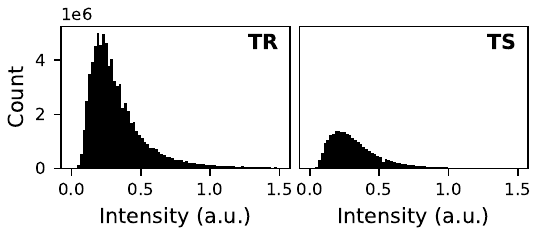}}
    \caption{Pixel intensity histograms for the training dataset \textbf{(TR)} and testing dataset \textbf{(TS)}. Pixel counts are binned as a function of normalized intensity of the photoacoustic signal across all imaging wavelengths used. In other words, each point represents a single $p(\mathbf{r}_i, \lambda_l)$ from either the training set or testing set. }
    \label{fig:distr}
\end{figure}

The contrast agents of interest in the \textit{in vivo} images were oxygenated and deoxygenated hemoglobin, therefore $N=2$. None of the images had any exogenous chromophores used. The tabulated absorption spectra were taken from \cite{prahlTabulatedMolarExtinction1998a}. A total of $5 \times 10^5$ sPA pixels were used for training and $1.5 \times 10^5$ for evaluating performance. The distributions of the training and testing sets are shown in Fig. \ref{fig:distr}. 

\subsection{Data Pre-Processing}

\begin{figure}[htb]
    \centerline{\includegraphics[width=\columnwidth]{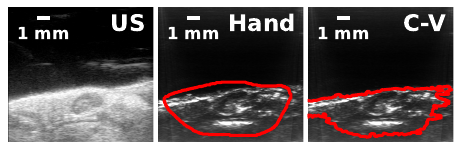}}
    \caption{Example highlighting segmentation of a sPA image using Chan-Vese active contours. \textbf{(US)} Ultrasound image of a mouse lymph node. \textbf{(Hand)} Initial hand-drawn segmentation of the associated sPA image before the Chan-Vese active contour method was applied. \textbf{(C-V)} Evolved segmentation of the sPA image using the Chan-Vese active contour method initiated from the \textbf{(Hand)} segmentation. Note the exclusion of the background pixels and the inclusion of more of the biologically relevant foreground.}
    \label{fig:seg}
\end{figure}

The sPA images were pre-processed by a segmentation procedure to enhance the quality of the training and testing datasets. The process of segmenting relevant pixels was two-fold. First, a region-of-interest (ROI) was drawn based on the B-Mode ultrasound and sPA image of a mouse lymph node. The rough manual segmentation was then fine-tuned by the Chan-Vese active contour method \cite{chanActiveContoursEdges2001a}. By employing an active-contour method, the ROI is composed primarily of biologically relevant foreground pixels and excludes background pixels. Fig. \ref{fig:seg} demonstrates the segmentation procedure on one of the \textit{in vivo} mouse lymph node sPA images. The segmentation procedure was applied to all fourteen sPA images used.\\

\subsection{Cost Functions and Training Procedure}
\label{sec:costdef}

SPOI-AE was trained using the Adam optimizer \cite{kingmaAdamMethodStochastic2017a}. Calculations were done in using Python 3.10.9 and PyTorch 1.13 on a computer equipped with an Intel i7-6700k CPU and two NVIDIA GTX 1080 GPUs. The implementation code is available at \cite{termartirosyanCodeOpticalInversion2023a}. A cross-validation procedure was employed to select the number of nodes per neural network layer for $\mu_a$-Net and $\mu_s'$-Net. The selected layer dimensions are outlined in Fig. \ref{fig:deepae}.

Traditionally, autoencoders are trained using mean-square-error (MSE) as the cost function \cite{ngSparseAutoencoder2011a}, defined as

\begin{equation}
\label{eq:mse}
\mathrm{MSE}\!\left(\mathbf{P},\,\widehat{\mathbf{P}}\right) = \frac{1}{I} \sum_{i=1}^{I} \left\|\mathbf{p}_i - \widehat{\mathbf{p}}_i\right\|_2^2.
\end{equation}

\noindent However, it has been shown that the spectral-angular-distance (SAD) is a useful auxiliary loss function to improve performance when fitting spectroscopic data \cite{ozkanEndNetSparseAutoEncoder2019a}. The mean-SAD (MSAD) function---defined in \eqref{eq:msad}---was used in tandem with MSE as the cost function used to train SPOI-AE. The combined cost function is defined in \eqref{eq:loss}.
 
\begin{equation}
\label{eq:msad}
\begin{aligned}
&\mathrm{SAD}\!\left(\mathbf{p}, \widehat{\mathbf{p}}\right) = \frac{2}{\pi} \cos^{-1}{\left\{\frac{\left\langle \mathbf{p}, \widehat{\mathbf{p}} \right\rangle}{\left\|\mathbf{p}\right\|_2 \left\|\widehat{\mathbf{p}}\right\|_2 + \texttt{EPSILON}}\right\}}
\\
&\mathrm{MSAD}\!\left(\mathbf{P},\,\widehat{\mathbf{P}}\right) = \frac{1}{I}\sum_{i=1}^{I} \mathrm{SAD}\!\left(\mathbf{p}_i,\,\widehat{\mathbf{p}}_i\right) 
\end{aligned}
\end{equation}

\begin{equation}
\label{eq:loss}
\mathcal{L}\!\left(\mathbf{P},\,\widehat{\mathbf{P}}\right) = \alpha\, \mathrm{MSE}\!\left(\mathbf{P},\,\widehat{\mathbf{P}}\right) + \beta\,\mathrm{MSAD}\!\left(\mathbf{P},\,\widehat{\mathbf{P}}\right)
\end{equation}

The hyperparameters $\alpha$ and $\beta$ in \eqref{eq:loss} are constant multipliers which balance out the contributions from MSE and MSAD. Based on empirical results, $\alpha$ was set to $10^2$ and $\beta$ to $5$. The MSAD component can also be disabled by setting $\beta$ to zero.

\subsection{Evaluation Metrics}
\label{sec:eval}

% \brown{We need a section for \textbf{Evaluation Metrics}.} \red{Done.}

In addition to the cost functions detailed in Section \ref{sec:costdef}, there are several evaluation metrics that are used to gauge model performance quantitatively and qualitatively. The metrics used primarily focus on measuring sPA pixel reconstruction accuracy across all wavelengths. Specifically,  reconstruction accuracy was quantified using wavelength dependent goodness-of-fit (GoF). Wavelength dependent GoF was calculated by finding the coefficient of determination ($R^2$) for each wavelength. This if formulated as

\begin{equation}
\label{eq:r2}
\begin{aligned}
% \xoverline{p}(\lambda_l) &= \frac{1}{I}\sum_{i=1}^{I} p_i(\lambda_l) \\
R^2(\lambda_l) &= 1 - \frac{\sum_{i=1}^{I}\left(p_i(\lambda_l) - \widehat{p}_i(\lambda_l) \right)^2}{\sum_{i=1}^{I}\left(p_i(\lambda_l) - \mathrm{E}_i\left[\mathbf{p}(\lambda_l)\right] \right)^2} \\ 
&= 1 - \mathrm{FVU}(\lambda_l)
\end{aligned}
\end{equation}

\noindent where the term $\mathrm{FVU}(\lambda_l)$ is the fraction of variance unexplained for the wavelength $\lambda_l$. A high-performing spectral unmixing algorithm will minimize $\mathrm{FVU}(\lambda_l)$ consistently across all wavelengths. In other words, it is desirable that an algorithm has a high average $R^2$ with little variation  across wavelengths. 

It is also possible to evaluate reconstruction accuracy by qualitatively comparing the average sPA spectra of the input data to the various reconstructions. The average spectra of a reconstruction or the original is found as 

\begin{equation}
\label{eq:avgsp}
\xoverline{p}(\lambda_l) = \mathrm{E}_i \left[\mathbf{p}(\lambda_l)\right]
\end{equation}

\noindent evaluated at all $\lambda_l$. Goodness of fit can by gauged by plotting plotting $\xoverline{p}(\lambda_l)$ for all tested algorithms and qualitatively comparing those spectra against the average input sPA spectrum. 

\subsection{Ground Truth Validation}
\label{sec:gtintro}

In order to interrogate the validity of any spectral unmixing method, it is necessary to compare estimated chromophore concentrations against known ground truth values. In this study, a computer-simulated phantom featuring three blood inclusions at various oxygenation levels embedded one centimeter deep serves as the ground truth. Optical absorption and reduced scattering coefficients were defined according to vessel oxygenation and typical properties, respectively. 

\begin{figure}[htb]
    \centering
    \includegraphics[width=\columnwidth]{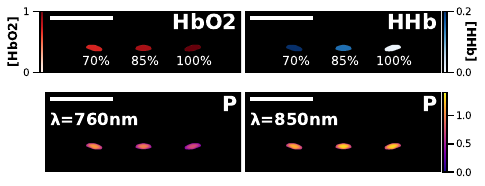}
    \caption{Ground-truth phantom oxygenated hemoglobin and deoxygenated hemoglobin concentration images (\textbf{(HbO2)} and \textbf{(HHb)}, respectively), with oxygenation of each inclusion annotated. Blood inclusions are embedded one centimeter deep into the simulation volume. Blood concentrations are ultimately mapped to initial pressure images, and two example wavelengths ($\lambda$=760nm and $\lambda$=850nm) are plotted in \textbf{P}.  Scalebars represent five millimeters.}
    \label{fig:gtdef}
\end{figure}

Using the generated optical coefficients, an optical simulation was performed using Monte Carlo eXtreme \cite{yanGraphicsprocessingunitacceleratedMonteCarlo2022} and an acoustic simulation was performed using the k-Wave toolbox \cite{treebyKWaveMATLABToolbox2010a}, both using MATLAB R2023a (MathWorks Corporation, Natick, MA, USA). Fig. \ref{fig:gtdef} presents the oxyhemoglobin and deoxyhemoglobin concentration maps of the simulated phantom, as well as two example reconstructed sPA wavelength slices of the full range (680--970nm spaced 2nm apart).   

\section{Results}

\subsection{Controls and Variations}

SPOI-AE's comparative performance was assessed against two linear alternatives. The first alternative is nonlinear-least-squares computed using tabulated absorption spectra from literature (Lit. NLS). The second was nonnegative-matrix-factorization (NMF) initialized with the tabulated absorption spectra from literature \cite{cichockiFastLocalAlgorithms2009a, fevotteAlgorithmsNonnegativeMatrix2011a}. Together, Lit. NLS and NMF represent a control to assess the improvement of the SPOI-AE against conventional spectral unmixing methods. It is worth noting that the latent space of both Lit. NLS and NMF is solely the relative chromophore concentrations, not the optical parameters of the tissue. 

SPOI-AE was also evaluated with different hyperparameter settings. Namely, performance was assessed with the MSAD cost-function component enabled (i.e., $\beta=5$) and disabled ($\beta=0$). Moreover, the performance cost of disabling the absorption spectra refining procedure was examined at both values of $\beta$. By comparing SPOI-AE against its variations, design choices are justified quantitatively. 
% \brown{Perhaps this paragraph could be moved to the end of \textbf{Methods} with a little bit more details on evaluation metrics?} 

\subsection{Summary Performance}

To get a general sense of algorithm performance, all SPOI-AE variants as well as Lit. NLS and NMF were tested using the cost metrics introduced in Section \ref{sec:costdef}. The results are presented in Table \ref{tab:results}, with the highest performing algorithm marked in \textbf{bold}. 

\begin{table}[htb]
    \caption{\normalsize Testing Set Performance}
    \label{tab:results}
    \centering
    \normalsize
    \begin{tabular}{\? c | c c \? c c \?}
        \toprule
        \multicolumn{3}{\?c\?}{Algorithm} & {MSE} & {MSAD} \\
        \midrule
        \multicolumn{3}{\?c\?}{Lit. NLS} & 0.0270 & 0.194 \\
        \multicolumn{3}{\?c\?}{NMF} & 0.0259 & 0.198  \\
        \multicolumn{3}{\?c\?}{SPOI-AE Variants} & & \\
        \cline{2-3}

        {     } & Adjusted $\mathbf{E}$ & $\beta=5$ & \B 0.0114 & \B 0.096  \\
        {     } & Adjusted $\mathbf{E}$ & $\beta=0$ &    0.0124 &    0.116  \\
        {     } & Fixed $\mathbf{E}$   & $\beta=5$ &    0.0126 &    0.100 \\
        {     } & Fixed $\mathbf{E}$   & $\beta=0$ &    0.0140 &    0.122  \\
        \bottomrule
    \end{tabular}
\end{table}

From Table \ref{tab:results}, SPOI-AE with adjusted absorption spectra and $\beta=5$ outperformed the other variations and controls with respect to MSE and MSAD. The summary statistics collected indicate that factoring MSAD into the training cost improves the MSE, regardless of if $\mathbf{E}$ was adjusted or not. Refining $\mathbf{E}$ improved reconstruction accuracy whether or not MSAD was used as a component of the cost function. This indicates that refining $\mathbf{E}$ might further compensate for confounding factors. 

\subsection{Wavelength Dependent Reconstruction}

\begin{figure}[htb]
    \centerline{\includegraphics[width=\columnwidth]{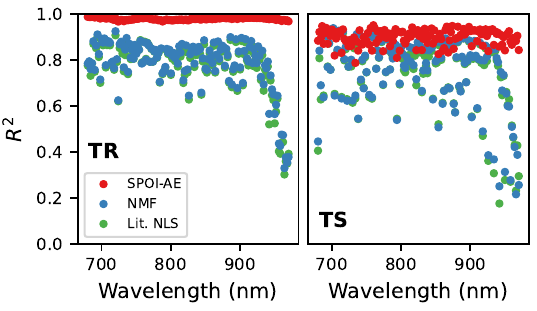}}
    \caption{$R^2$ results for the training set \textbf{(TR)} and the testing set \textbf{(TS)} as a function of imaging wavelength in nm.}
    \label{fig:r2}
\end{figure}

As described in Section \ref{sec:eval}, it is important to evaluate the candidate algorithms for wavelenght dependent goodness-of-fit. Using \eqref{eq:r2}, the $R^2$ values were calculated for Lit. NLS and NMF as well as SPOI-AE with adjusted spectra and $\beta=5$. $R^2$ was calculated for both the training and testing sets. The results are presented in Fig. \ref{fig:r2} and Table \ref{tab:r2}. In Table \ref{tab:r2}, the highest performing algorithm for each metric is marked in \textbf{bold}.

\begin{table}[htb]
    \caption{\normalsize Testing Set Wavelength Dependent GoF}
    \label{tab:r2}
    \centering
    \normalsize
    \begin{tabular}{\? c | c c \? c c \?}
        \toprule
        \multicolumn{3}{\?c\?}{Algorithm} & {$\mathrm{E}[R^2]$} & {$\sqrt{\mathrm{Var}[R^2]}$} \\
        \midrule
        \multicolumn{3}{\?c\?}{Lit. NLS} & 0.741 & 0.1559 \\
        \multicolumn{3}{\?c\?}{NMF} & 0.750 & 0.156 \\
        \multicolumn{3}{\?c\?}{SPOI-AE Variants} & & \\
        \cline{2-3}

        {     } & Adjusted $\mathbf{E}$ & $\beta=5$ & \B 0.897 & \B 0.0331 \\
        {     } & Adjusted $\mathbf{E}$ & $\beta=0$ &    0.889 &    0.0443 \\
        {     } & Fixed $\mathbf{E}$   & $\beta=5$ &    0.887 &    0.0412 \\
        {     } & Fixed $\mathbf{E}$   & $\beta=0$ &    0.875 &    0.0478 \\
        \bottomrule
    \end{tabular}
\end{table}

Fig. \ref{fig:r2} and Table \ref{tab:r2} present corroborating results, namely that SPOI-AE better reconstructs an input sPA pixel than the alternative methods on average. The wavelength dependent GoF results echo the summary performance statistics outlined in Table \ref{tab:results}, with the deep autoencoder variants outperforming Lit. NLS and NMF. 

\begin{figure}[htb]
    \centerline{\includegraphics[width=\columnwidth]{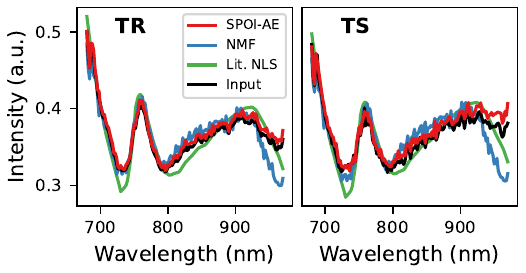}}
    \caption{Average sPA pixel spectrum reconstruction versus average input sPA spectrum. SPOI-AE \textbf{(red)} better reconstructs the input than NMF \textbf{(blue)} and Lit. NLS \textbf{(green)} for the average training pixel \textbf{(TR)}. This performance translates to the average testing pixel \textbf{(TS)}. For both plots, normalized spectra is drawn as a function of imaging wavelength in nm. }
    \label{fig:avgsp}
\end{figure}

Reconstruction accuracy can also be gauged qualitatively by examining the average reconstructed spectrum versus the average pixel. Equation \eqref{eq:avgsp} details how average spectra can be computed for the input and all tested wavelength. Moreover, Section \ref{sec:eval} details how average reconstructed spectra can be compared to the average input spectra to gauge performance qualitatively. The average spectra are presented in  Fig. \ref{fig:avgsp}. 

Compared to NMF and Lit. NLS, SPOI-AE is better able to follow the shape of the average training and testing input spectra, indicating superior reconstruction performance. The improved performance is likely thanks to SPOI-AE's intermediate optical inversion and use of the MSAD cost function, mitigating the effects of spectral coloring and parasitic absorption at longer laser wavelengths. 

\subsection{Optical Inversion}

As detailed in Section \ref{sec:nndesign}, SPOI-AE generates estimates for the photoacoustic optical parameters. Due to the unlabeled nature of the dataset, there is no way of assessing the accuracy of this estimate, but it is possible to evaluate it for qualitative physical coherence. For example, on rule-of-thumb regarding photoacoustic scattering and absorption is that the reduced scattering coefficient $\mu_s'$ should be approximately two orders of magnitude greater than the absorption coefficient $\mu_a$ \cite{petersOpticalPropertiesNormal1990b}. 

% \red{More sanity checks needed to complete section, I think? However, not sure what those would be.} 

% \brown{Could labeled phantoms quantitatively evaluate this estimate?} \red{Yes, but definitely out-of-scope for this paper.}

\begin{figure}[t]%{0.495\linewidth}
    \centerline{\includegraphics[width=\columnwidth]{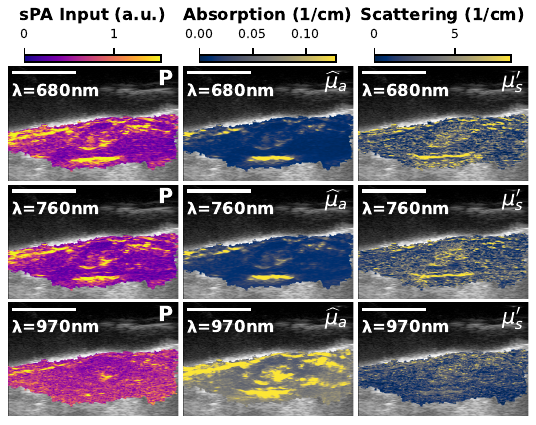}}
    \caption{Optical parameters for the \textit{in vivo} mouse lymph node image introduced in Fig. \ref{fig:seg} at imaging wavelengths $\lambda$=680 nm, $\lambda$=760 nm, and $\lambda$=970 nm. \textbf{(P)} Input photoacoustic images captured at the indicated laser wavelength in normalized magnitude. \textbf{($\widehat{\boldsymbol\mu}_a$)} Reconstructed absorption coefficient image estimated using SPOI-AE for the indicated laser wavelength estimated using \eqref{eq:muahat}. \textbf{($\boldsymbol\mu_s'$)} Reduced scattering coefficient image estimated using SPOI-AE at the indicated laser wavelength found using \eqref{eq:nets}. Scalebars indicate five millimeters.}
    \label{fig:mua_mus}
\end{figure}

Fig. \ref{fig:mua_mus} displays the optical parameters for the \textit{in vivo} lymph node image introduced in Fig. \ref{fig:seg}. Qualitative analysis suggests that these estimates are physically coherent. Moreover, the $\widehat{\mu}_a$ and $\mu_s'$ images in Fig. \ref{fig:mua_mus} indicate where absorbing and scattering species are located spatially in the image, allowing for the interpretation of SPOI-AE's behavior. 

\subsection{Spectral Unmixing}
\label{sec:unmixing}

\begin{figure}[htb]
    \centerline{\includegraphics[width=\columnwidth]{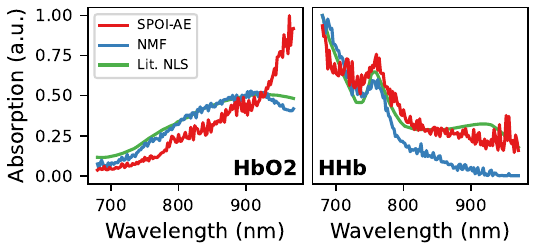}}
    \caption{Relative absorption spectra for oxygenated hemoglobin \textbf{(HbO2)} and deoxygenated hemoglobin \textbf{(HHb)} as a function of imaging wavelength in nm. }
    \label{fig:spectra}
\end{figure}

As described in Section \ref{sec:latent}, the absorption spectra $\mathbf{E}$ can be a tunable  parameter of the machine learning process. Absorption spectra estimated by SPOI-AE are compared against alternatives in Fig. \ref{fig:spectra}. The spectra estimates generated by SPOI-AE and NMF deviate from the literature spectra in response to both methods' respective cost functions. Note that since both SPOI-AE and NMF are trained in a self-supervised manner, it is impossible to know if the modified spectra describe the true absorption spectra of HbO2 or HHb. 

The absorption spectra in Fig. \ref{fig:spectra} were used to perform spectral unmixing. In the case of SPOI-AE, spectral unmixing was performed using the mechanism in \eqref{eq:cvec}. For both Lit. NLS and NMF, spectral unmixing was performed according to the respective outlines in Section \ref{sec:linsu}. 

By unmixing HbO2 and HHb using either Lit. NLS, NMF, or SPOI-AE, it is possible to estimate the oxygen saturation with the formula

\begin{equation}
\label{eq:so2}
\mathrm{SO2}(\mathbf{r}_i) = \frac{c_{\mathrm{HbO2}}(\mathbf{r}_i)}{c_{\mathrm{HbO2}}(\mathbf{r}_i) + c_{\mathrm{HHb}}(\mathbf{r}_i)} \times 100\%.
\end{equation}

\begin{figure}[thb]%{0.495\linewidth}
    \centerline{\includegraphics[width=\columnwidth]{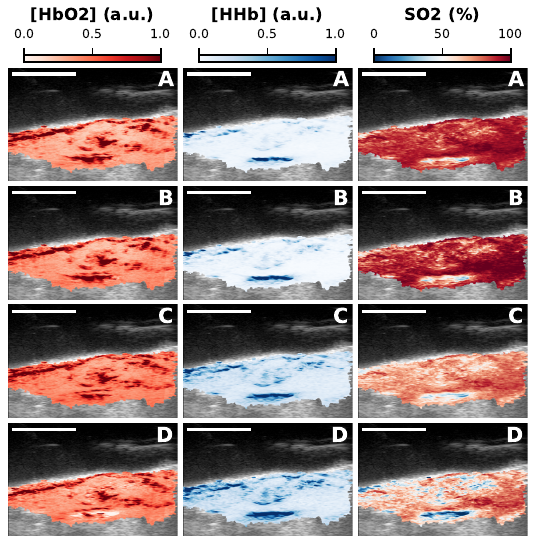}}
    \caption{Relative HbO2 and HHb concentration maps as well as SO2 percentage maps calculated using \textbf{(A)} SPOI-AE with adjusted absorption spectra, \textbf{(B)} SPOI-AE with fixed absorption spectra from literature, \textbf{(C)} NMF, and \textbf{(D)} Lit. NLS. Scalebars indicate five millimeters.}
    \label{fig:unmixing}
\end{figure}

\begin{figure}[!htb]
    \centerline{\includegraphics[width=\columnwidth]{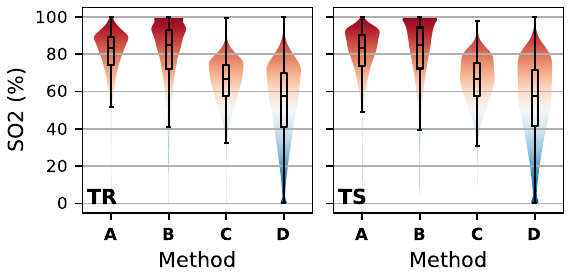}}
    \caption{SO2 distributions for the training \textbf{(TR)} and testing \textbf{(TS)} datasets as well as for each unmixing method investigated herein, namely: \textbf{(A)} SPOI-AE with adjusted absorption spectra, \textbf{(B)} SPOI-AE with fixed absorption spectra from literature, \textbf{(C)} NMF, and \textbf{(D)} NLS with literature spectra. The SO2 values provided by NMF and Lit. NLS are consistently lower than the healthy range.}
    \label{fig:so2distr}
\end{figure}

Relative chromophore concentration maps for both HbO2 and HHb as well as SO2 percentages are shown in Fig. \ref{fig:unmixing}. We can observe that the SO2 values estimated using SPOI-AE are distributed closer to 100\% oxygenation than those rendered by NMF or Lit. NLS. SO2 values in the neighborhood of 90\% to 100\% are expected for healthy mouse tissue \cite{flecknellLaboratoryAnimalAnaesthesia2015a}. We can inspect the specific oxygenation distributions across both the training and testing datasets in Fig. \ref{fig:so2distr} and see that SPOI-AE yields more reasonable SO2\% distributions versus Lit. NLS and NMF.

\subsection{Ground Truth Validation}

\begin{figure}[htb]%{0.495\linewidth}
    \centerline{\includegraphics[width=\columnwidth]{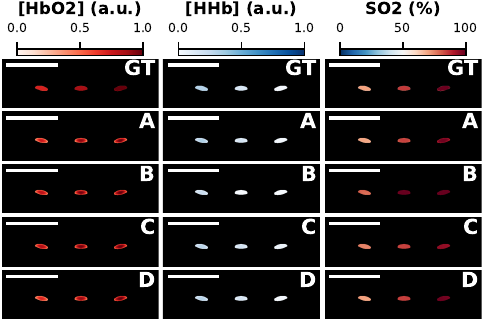}}
     \caption{Relative HbO2 and HHb concentration maps and SO2 percentage maps of the ground truth simulation phantom \textbf{(GT)} calculated using \textbf{(A)} SPOI-AE with adjusted absorption spectra, \textbf{(B)} SPOI-AE with fixed absorption spectra from literature, \textbf{(C)} NMF, and \textbf{(D)} Lit. NLS. Scalebars indicate five millimeters. }
    \label{fig:gtunmixing}
\end{figure}

To verify the accuracy of SPOI-AE's spectral unmixing, the simulated blood vessel sPA image from Section \ref{sec:gtintro} was used. The estimated chromophore concentrations maps as well as SO2 estimates found using the various methods investigated herein are shown in Fig. \ref{fig:gtunmixing} versus the ground-truth. 

Blood unmixing accuracy by method was found by computing the mean-absolute error of SO2 estimates in the vessel regions. The SO2 in percentage-points (pp) are reported in Table \ref{tab:so2_mae}.

\begin{table}[htb]
    \caption{\normalsize Simulated Phantom SO2 Unmixing Error}
    \label{tab:so2_mae}
    \centering
    \normalsize
    \begin{tabular}{\? c | c c \? c \?}
        \toprule
        \multicolumn{3}{\?c\?}{Algorithm} & {$\mathrm{MAE}\left[\mathrm{SO2}\right]$ (pp)} \\
        \midrule
        \multicolumn{3}{\?c\?}{Lit. NLS} & \B 0.657 \\
        \multicolumn{3}{\?c\?}{NMF} & 3.64 \\
        \multicolumn{3}{\?c\?}{SPOI-AE Variants} & \\
        \cline{2-3}
        {     } & Adjusted $\mathbf{E}$ & $\beta=5$ & 2.63  \\
        {     } & Adjusted $\mathbf{E}$ & $\beta=0$ & 4.18 \\
        {     } & Fixed $\mathbf{E}$   & $\beta=5$ & 6.59 \\
        {     } & Fixed $\mathbf{E}$   & $\beta=0$ & 7.79  \\
        \bottomrule
    \end{tabular}
\end{table}

% \newpage
\section{Discussions and Conclusions}

Optical inversion and spectral unmixing of spectroscopic photoacoustic images is an intractable problem due to nonlinearity and ill-posedness \cite{coxChallengesQuantitativePhotoacoustic2009a}. Conventional spectral unmixing methods---namely, nonnegative-least-squares and nonnegative-matrix-factorization---make the fundamentally flawed assumption that a sPA image pixel can be described as a low-rank linear product between absorption spectra and relative chromophore concentrations \cite{saratoonGradientbasedMethodQuantitative2013a}. The presented ``SPOI-AE'' model addresses this flaw by performing spectral unmixing on estimated optical parameters rather than the input sPA images. 

SPOI-AE provides a framework for estimating the scattering and absorption coefficients of a sPA image, solving the photoacoustic optical inverse problem. SPOI-AE can optionally modify absorption spectra to fit the input sPA data better. While trained in a self-supervised manner, SPOI-AE is able to estimate physically and biologically coherent estimates for the optical parameters thanks to using the photoacoustic optical forward problem as the autoencoder decoding mechanism. SPOI-AE's unmixing ability was further verified using a simulated ground truth phantom. 

Quantitative and qualitative metrics demonstrate that SPOI-AE better reconstructs input sPA pixels than alternative methods. This improved performance is likely thanks to using the photoacoustic optical forward problem as the unmixing mechanism, tying SPOI-AE to the constraint of physical optical transport. The improved wavelength dependent goodness-of-fit can also likely be attributed to using the mean-spectral-angular-distanjce as an auxiliary cost metric, ensuring consistent performance across all wavelengths. 

Testing using a simulated phantom with a swatch of oxygenated inclusions revealed that SPOI-AE variants with adjusted $\mathbf{E}$ performed comparably with NMF, with fixed $\mathbf{E}$ variants being slightly less accurate. Crucially, the SPOI-AE variant with adjusted extinction spectra and a MSAD error term exhibited 2.63 pp of error, positioning this neural network as a capable tool for discriminating hypoxia or hyperoxia from normoxia. Future work incorporating more data and variations to network architecture could ellicit similar performance from the fixed extinction SPOI-AE variants. 

Moreover, future studies could look at including more simulated and real-world ground truth data in testing SPOI-AE. The SPOI-AE architecture can be easily modified to employ a semi-supervised training scheme by including some labeled sPA pixels into the dataset. In future studies, \textit{ex vivo} mouse experiments with inserted tubes of blood with known SO2 will enable such an extension. 

SPOI-AE can be modified in the future to use a variational estimation scheme, such as a variational-autoencoder (VAE) or a disentanglement-VAE ($\beta$-VAE) as the model architecture \cite{kingmaAutoEncodingVariationalBayes2022a, higginsBetaVAELearningBasic2016a}. By employing a VAE or $\beta$-VAE, the SPOI-AE would be able to learn statistical distributions for the optical parameters and chromophore concentrations, unlocking uncertainty quantification.

In this study, we proposed SPOI-AE: an effective tool for optically inverting spectroscopic photoacoustic images of mouse lymph nodes. SPOI-AE is demonstrably better than linear alternatives at reconstructing input sPA pixels from a latent space. Despite being trained in a self-supervised manner with no ground truth images, the latent space estimates are logical, and can be further improved with an expanded, semi-labeled dataset. Ultimately, the SPOI-AE architecture is a promising framework for the optical inversion and spectral unmixing of spectroscopic photoacoustic images using physics-informed neural networks.  

\section{Acknowledgment}

The authors would like to thank their colleagues in the Ultrasound Imaging and Therapeutics Research Laboratory for their input, advice, and support on this project. 

% \clearpage
\bibliographystyle{IEEEtran} 
\bibliography{preprint_refs}

\end{document}